\documentclass[11pt, letterpaper]{nyu}
\usepackage{amsmath,amsfonts,bm}

\def\eqref#1{equation~\ref{#1}}

\def\1{\bm{1}}

\DeclareMathAlphabet{\mathsfit}{\encodingdefault}{\sfdefault}{m}{sl}
\SetMathAlphabet{\mathsfit}{bold}{\encodingdefault}{\sfdefault}{bx}{n}

\newcommand{\website}{\href{https://dynamem.github.io}{dynamem.github.io}}
\websitedef{\website}
\usepackage{subcaption} %
\usepackage{times} %
\usepackage{amssymb}  %
\usepackage[skip=4pt,font=small,labelfont=bf]{caption}
\usepackage[sort, square, compress]{natbib}
\setcitestyle{numbers}
\setcitestyle{citesep={,}}
\usepackage{hyperref}[citecolor=lightblue]
\usepackage{xcolor}  %
\definecolor{customcitecolor}{RGB}{125,23,217}  %
\definecolor{customlinkcolor}{RGB}{176,87,228}
\hypersetup{
    colorlinks=true,
    linkcolor=customcitecolor,
    filecolor=customcitecolor,      
    citecolor=customcitecolor,
    urlcolor=customcitecolor,
}
\usepackage{graphicx}
\usepackage{enumitem}
\usepackage{amsmath}
\usepackage{algpseudocode}
\usepackage{algorithm}
\usepackage[algo2e]{algorithm2e}
\usepackage{multirow}
\usepackage[normalem]{ulem}
\usepackage{booktabs}
\usepackage{verbatim}

\newcommand{\MODEL}[0]{DynaMem}
\newcommand{\BENCH}[0]{DynaBench}
\newcommand{\ENVNUM}[0]{9}

\newcommand{\mysection}[1]{%
  \vspace{0.25\baselineskip}
  \par\noindent\textbf{#1}:%
}
\newcommand{\xxnote}[3]{}
\ifx\hidenotes\undefined
  \renewcommand{\xxnote}[3]{\color{#2}{#1: #3}}
\fi

\title{
\MODEL{}: Online Dynamic Spatio-Semantic Memory for Open World Mobile Manipulation
}

\author[1]{Peiqi Liu}
\author[1]{Zhanqiu Guo}
\author[1]{Mohit Warke}
\author[1, 3]{Soumith Chintala}
\author[2]{Chris Paxton}
\author[1]{Nur Muhammad Mahi Shafiullah$^{*}$}
\author[1]{Lerrel Pinto$^{*}$}
\affil[1]{New York University}
\affil[2]{Hello Robot Inc.}
\affil[3]{Meta Inc.}

\correspondingauthor{ \href{mailto:pl2285@nyu.edu}{pl2285@nyu.edu}, \href{mailto:mahi@cs.nyu.edu}{mahi@cs.nyu.edu}. $(^{*})$~denotes equal advising.}

\begin{document}

\begin{abstract}
    Significant progress has been made in open-vocabulary mobile manipulation, where the goal is for a robot to perform tasks in any environment given a natural language description.
    However, most current systems assume a static environment, which %
    limits the system's applicability in real-world scenarios where environments frequently change due to human intervention or the robot's own actions.
    In this work, we present~\MODEL{}, a new approach to open-world mobile manipulation that uses a dynamic spatio-semantic memory to represent a robot's environment.
    ~\MODEL{} constructs a 3D data structure to maintain a dynamic memory of point clouds, and answers open-vocabulary object localization queries using multimodal LLMs or open-vocabulary features generated by state-of-the-art vision-language models.
    Powered by~\MODEL{}, our robots can explore novel environments, search for objects not found in memory, and continuously update the memory as objects move, appear, or disappear in the scene.
    We run extensive experiments on the Stretch SE3 robots in three real and nine offline scenes, and achieve an average pick-and-drop success rate of 70\% on non-stationary objects, which is more than a $\mathbf{2\times}$ improvement over state-of-the-art static systems.
\end{abstract}

\maketitle
\begin{figure}[h!]
    \centering
    \includegraphics[width=\textwidth]{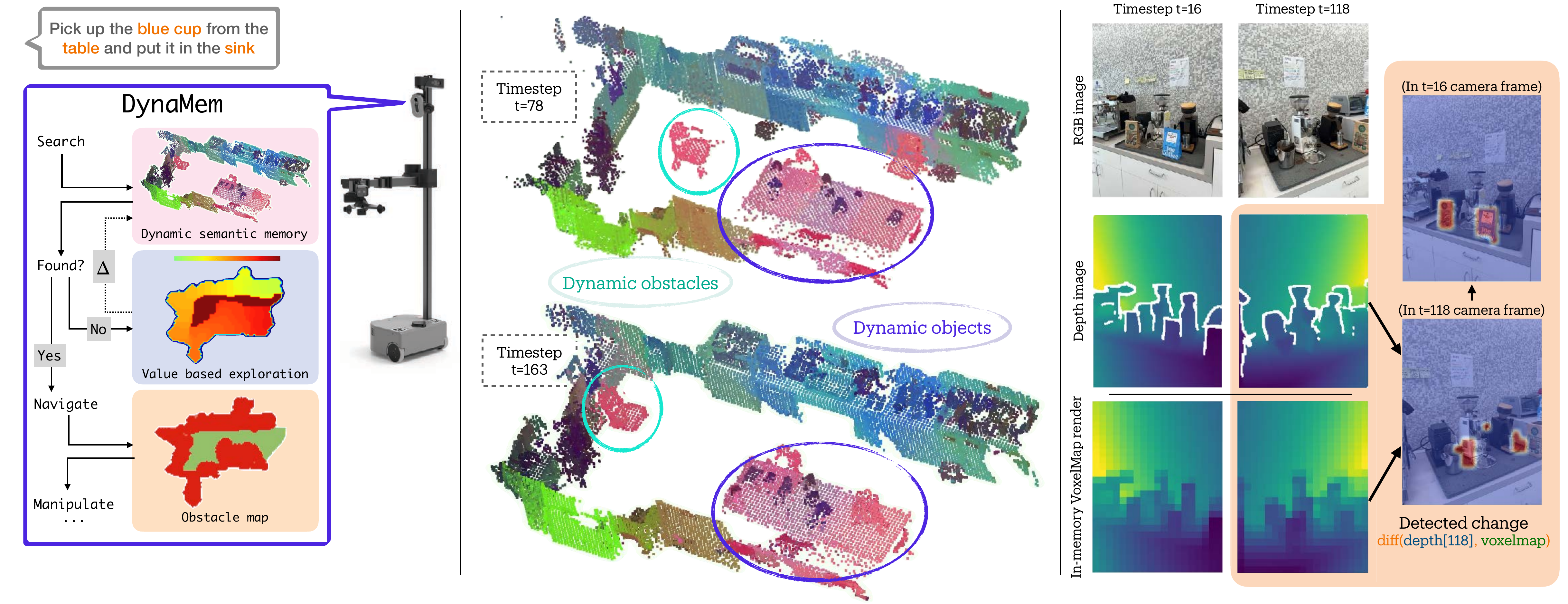}
    \caption{An illustration of how our online dynamic spatio-semantic memory \MODEL{} responds to open vocabulary queries in a dynamic environment. During operation and exploration,~\MODEL{} keeps updating its semantic map in memory. \MODEL{} maintains a voxelized pointcloud representation of the environment, and updates with dynamic changes in the environment by adding and removing points.}
    \label{fig:teaser}
    \vspace{-2em}
\end{figure}

\section{Introduction}

Recent advances in robotics have made it possible to deploy robots in real world settings to tackle the open vocabulary mobile manipulation (OVMM) problem~\cite{homerobot}.
Here, the robots are tasked with navigating in unknown environments and interacting with objects following open vocabulary language instructions, such as ``Pick up \textbf{X} from \textbf{Y} and put it in \textbf{Z}'', where X, Y, and Z could be any object name or location.
The two most common approaches to tackling OVMM are using policies trained in simulation and deploying them in the real world~\cite{ehsani2023imitating,Ramrakhya_2023_CVPR,poliformer}, or training modular systems that combine open vocabulary navigation (OVN)~\cite{clipfields,conceptgraphs, clio, lerf} with different robot manipulation skills~\cite{okrobot,  geff, usa-net, goat, hovsg}.
Modular systems enjoy greater efficiency and success in real-world deployment~\cite{gervet2023navigating} as they can directly leverage advances in vision and language models~\cite{okrobot, goat}, and are able to handle more diverse and out-of-domain environments with no additional training.

However, as recent analysis has shown, the primary challenge in deploying modular OVMM is that limitations of a module propagate to the entire system~\cite{okrobot}. One key module in any OVMM system is the open vocabulary navigation (OVN) module responsible for navigating to goals in the environment.
While many such OVN systems have been proposed in the literature~\cite{homerobot, lerf, clipfields, usa-net, geff,okrobot, conceptgraphs, clio, goat, hovsg}, they share a common limitation: they assume static, unchanging environments.
Contrast this with the real world, where environments change and objects are moved by either robots or humans. %
Making such a restrictive assumption thus limits these systems' applicability in real-world settings.
The primary reason behind this assumption is the lack of an effective dynamic spatio-semantic memory that can adapt to both addition and removal of objects and obstacles in the environment online.

In this work, we propose a novel spatio-semantic memory architecture, Dynamic 3D Voxel Memory (\MODEL{}), that can adapt  online to changes in the environment. \MODEL{} maintains a voxelized pointcloud representation of an environment and adds or removes points as it observes the environment change. Additionally, it supports two different ways to query the memory with natural language: a vision-language model (VLM) featurized pointcloud, and a multimodal-LLM (mLLM) QA system.
Finally,~\MODEL{} enables efficient exploration in changing environments by offering a dynamic obstacle map and a value-based exploration map that can guide the robot to explore unseen, outdated, or query-relevant parts of the world.

We evaluate~\MODEL{} as a part of full open-vocabulary mobile manipulation stack in three real world environments with multiple rounds of changes and manipulating multiple non-stationary objects, improving the static baseline by more than $2\times$ (70\% vs. 30\%). Additionally, we identify an obstacle in efficiently developing dynamic spatio-semantic memory, namely the lack of dynamic benchmarks, since many OVN systems use static simulated environments~\cite{scanrefer,scannet} or static datasets~\cite{yadav2023habitat,baruch2021arkitscenes}. We address this by developing a new dynamic benchmark, \BENCH{}. It consists of \ENVNUM{} different environments,
each changing over time. We ablate our design choices in this benchmark.
To the best of our knowledge,~\MODEL{} is the first spatio-semantic memory structure supporting both adding and removing objects.

\section{Related Works}
\label{sec:related-work}

\subsection{Open Vocabulary Mobile Manipulation (OVMM)}
Navigating to arbitrary goals in open ended environments and manipulating them has become a key challenge in robotic manipulation~\cite{homerobotovmmchallenge2023,ASC2023}.
This line of query follows Open-Vocabulary Navigation systems~\cite{clipfields,vlmaps}, which builds upon prior object and point goal navigation literature~\cite{gervet2023navigating,majumdar2020improving,krantz2022instance,hahn2021no,NEURIPS2020_2c75cf26,9636743,Zhao_2021_ICCV,DBLP:journals/corr/abs-2006-13171,goat} which attempted navigation to points, or fixed set of objects and object categories.
OVMM is a naturally harder challenge as it requires an ability to handle arbitrary queries, and ``navigation to manipulation'' transfer -- which means unlike pure navigation, the robot needs to get close to the environment objective and obstacles.
In the OVMM challenge~\cite{homerobotovmmchallenge2023}, modular solutions such as~\cite{homerobot,uniteam,hovsg} outperformed the competition. More recently, OK-Robot~\cite{okrobot} performed extensive real-world evaluations of the challenges in OVMM and demonstrated a system that achieves 58.5\% success rate in static home environments. We extend this work by enabling manipulation in changing environments.

\subsection{Spatio-semantic Memory}
Early works in spatio-semantic memory~\cite{peter2012mapping,bowman2017probabilistic,zhang2018semantic,ma2017multi,chaplot2020object} created semantic maps for limited categories based on mostly ad-hoc deep neural networks.
Later work builds upon representations derived from pre-trained vision language models, such as~\cite{ha2022semantic,shafiullah2022clip,huang2023visual,chen2022nlmapsaycan,jatavallabhula2023conceptfusion,kerr2023lerf,conceptgraphs,clio}.
These works use a voxel map or neural feature field as their map representation.
Some recent models~\cite{ji2024graspsplats,shorinwa2024splat} have used Gaussian splats~\cite{kerbl3Dgaussians} to represent semantic memory for manipulation.
Most of these models show object localization in pre-mapped scenes, while CLIP-Fields~\cite{clipfields}, VLMaps~\cite{vlmaps}, and NLMap-SayCan~\cite{chen2022nlmapsaycan} show integration with real robots for indoor navigation tasks.
Some recent works~\cite{bolte2023usa,wang2023d,geff} extend this task to include an affordance model or manipulation primitives.
Our work builds upon the voxel map based spatio-semantic memory literature and extends them to dynamic environments where both objects and obstacles can change over time.
Concurrent to our work, DovSG~\cite{yan2024dynamicopenvocabulary3dscene} looks at dynamic semantic scene graphs. As scene graphs deal with an object level abstraction, DovSG needs to handle object merging, association, and deduplication explicitly, which are all handled implicitly in~\MODEL{}.

\subsection{Mapping and Navigating Dynamic Environments}
For robot navigation, Simultaneous Localization and Mapping (SLAM)~\cite{durrant2006simultaneous} methods are crucial.
However, practical SLAM instances based on voxels~\cite{song2024voxelnextfusion,shi2021rgb}, objects~\cite{mccormac2018fusion++,krishna20233ds}, landmark~\cite{bowman2017probabilistic,michael2022probabilistic}, NeRF~\cite{maggio2023loc,rosinol2023nerf}, and Gaussian splats~\cite{matsuki2024gaussian, yan2024gs} tend to make the simplifying assumption that the world is static.
Some sparse SLAM methods improve on dynamic environments by estimating underlying state~\cite{qiu2022airdos,cui2019sof,brasch2018semantic,yu2018ds,song2022dynavins,yu2021fusing,bescos2018dynaslam,khronos,virgolino2023visual} or explicitly modeling moving objects~\cite{bescos2021dynaslam,henein2020dynamic,henning2022bodyslam}.
Some methods also forego a map and rely on reactive policies to navigate dynamic environments~\cite{haviland2022holistic,ahn2022can,du2022bayesian,wong2022error,uppal2024spinsimultaneousperceptioninteraction}, although they generally tackle local movement and not global navigation.
Our work relies on SLAM systems that are stable under environment dynamics, and focuses on building a dynamic semantic memory based off of online exploration and observations.

\section{Method}
\label{sec:method}
In this section, we define our problem setup, and then describe our online, dynamic spatio-semantic memory for open world, open vocabulary mobile manipulation.

\subsection{Problem Statement}
\label{subsec:problem}

We create our algorithm,~\MODEL{}, to solve open vocabulary mobile manipulation (OVMM) problems in an open, constantly changing world.
The goal in OVMM is for a mobile robot to execute a series of manipulation commands given arbitrary language goals.
We assume the following requirements for the memory module for dynamic, online operation:

\begin{itemize}[leftmargin=1em]
    \item \textbf{Observations:} The mobile robot is equipped with an on-board RGB-D camera, and unlike prior work~\cite{okrobot}, doesn't start with a map of the environment. Rather, the robot explores the world and use the online observed sequence of posed RGB-D images to build its map.
    \item \textbf{Environment dynamism:} The environment can change without the knowledge of the robot. %
    \item \textbf{Localization queries:} Given a natural language query (i.e. "teddy bear"), the memory module has to return the 3D location of the object or determine that the object doesn't exist in the scene observed thus far.
    \item \textbf{Obstacle queries:} The memory module must determine whether a point in space is occupied by an obstacle. Both the location of the objects and obstacles can move, previous observations often contradict each other and must be resolved by the memory.
\end{itemize}

Note the significant upgrade in challenge in multiple facets compared to prior work~\cite{homerobot, lerf, clipfields, usa-net, geff,okrobot, conceptgraphs, clio, goat, hovsg}: almost no prior work dealing with open-vocabulary queries support dynamic environments with both addition and deletion, some assumes access to prior map data, and many don't handle \textit{negative} results, i.e. objects not found in memory, and instead return the best match.

\subsection{Dynamic 3D Voxel Map}
\label{subsec:updating}

\begin{figure}
\centering
\begin{subfigure}{.43\textwidth}
  \centering
    \includegraphics[width=\linewidth]{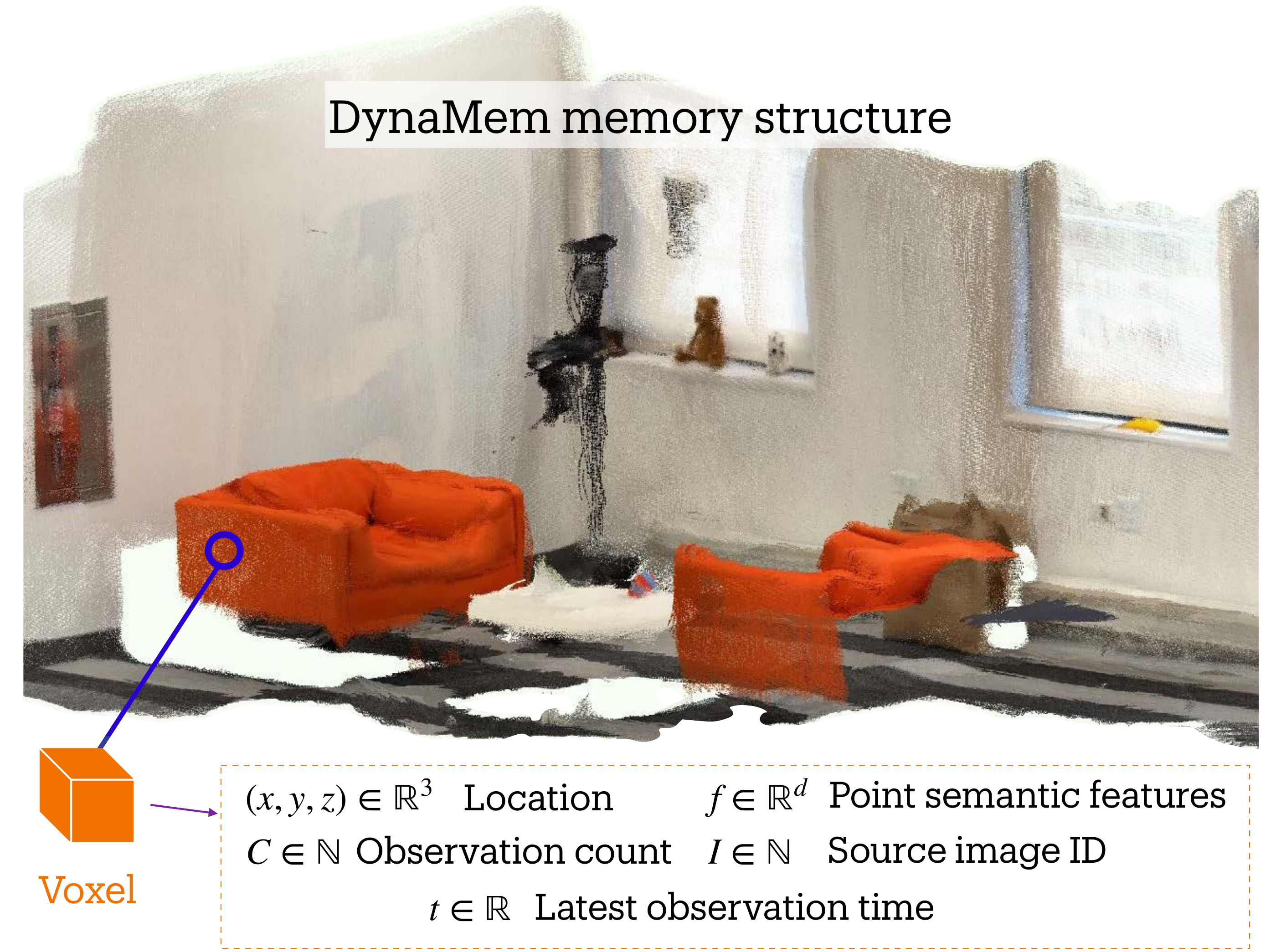}
\end{subfigure}%
\begin{subfigure}{.57\textwidth}
  \centering
    \includegraphics[width=\linewidth]{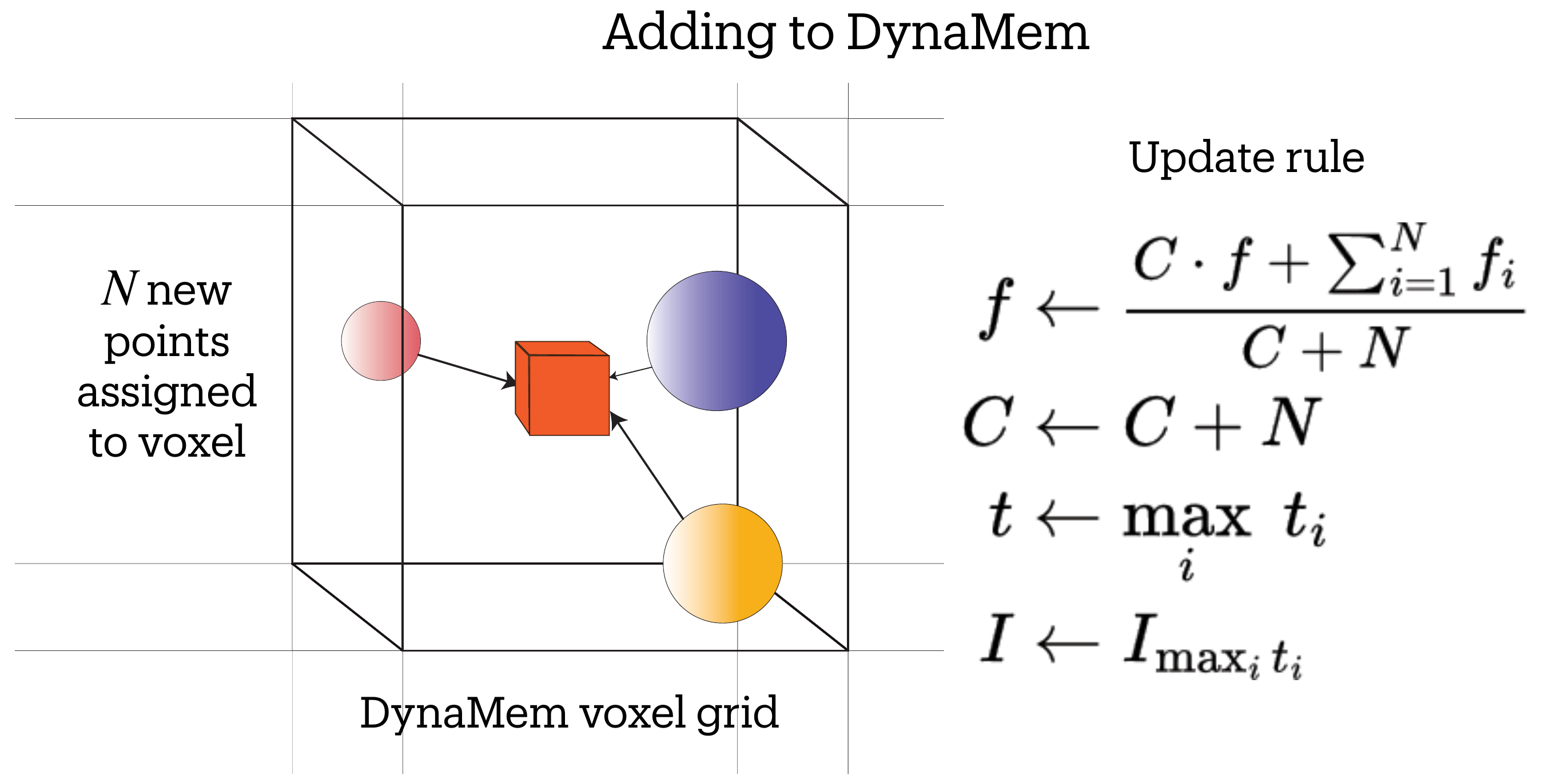}
\end{subfigure}
\caption{(Left) \MODEL{} keeps its memory stored in a sparse voxel grid with associated information at each voxel. (Right) Updating~\MODEL{} by adding new points to it, alongside the rules used to update the stored information.}
\label{fig:dynamem-structure}
\end{figure}

Our answer to the challenge posed in the Section~\ref{subsec:problem} is \MODEL{}.
\MODEL{} is an evolving sparse voxel map with associated information stored at each voxel, as shown in Figure~\ref{fig:dynamem-structure}.
In each non-empty voxel, alongside its 3D location $(x, y, z)$, we also store the observation count $C$ (how many times that voxel was observed), source image ID $I$ (which image the voxel was backprojected from), a high-dimensional semantic feature vector $f$ coming from a VLM like CLIP~\cite{clip} or SigLIP~\cite{siglip}, and the latest observation time, $t$, in seconds.

To make this data structure dynamic, we describe the process with which we add and update with new observations and remove outdated objects and associated voxels.

\mysection{Adding Points}
When the robot receives a new set of observations, i.e. RGB-D images with global poses, we convert them to 3D coordinates in a global reference frame, and generate a semantic feature vector for each point. The global coordinates are calculated from the global camera pose and the backprojected depth image using the known camera transformation matrix.
We calculate the point-wise image feature by first converting the images to object patches by using a segmentation model such as SAM-v2~\cite{sam2}, and then aggregating each patch feature over the output of a vision-language models like CLIP~\cite{clip} or SigLIP~\cite{siglip}.
For more details about image-to-feature vector mapping, we refer to earlier works~\cite{clipfields, okrobot, lerf}.
Once we have calculated the points and associated features, we cluster the new points and assign them to the nearest voxel grids.
In Figure~\ref{fig:dynamem-structure}, we show how each voxel's metadata is updated. The count keeps track of the total number of assigned points to each voxel grid, and the feature vector keeps track of the weighted average of all feature vectors assigned to that voxel.
Finally, the observation time and image ID are updated to keep track of the latest observation contributing to a particular voxel.
If a voxel was empty before assignment, we assume its count $C = 0$ and feature vector $f = \overrightarrow{0}$.

\mysection{Removing Points} When an object is moved or removed, its associated voxels in~\MODEL{} may get removed.
We use ray-casting to find the outdated voxels.
The operation follows a simple principle: if a voxel falls within the frustum between the camera plane and the associated view point cloud, that voxel must be unoccupied.
To reduce the impact of the depth noise at long range, we don't consider any pixel whose associated depth value is over 2m.
\begin{figure}
    \centering
    \includegraphics[width=\linewidth]{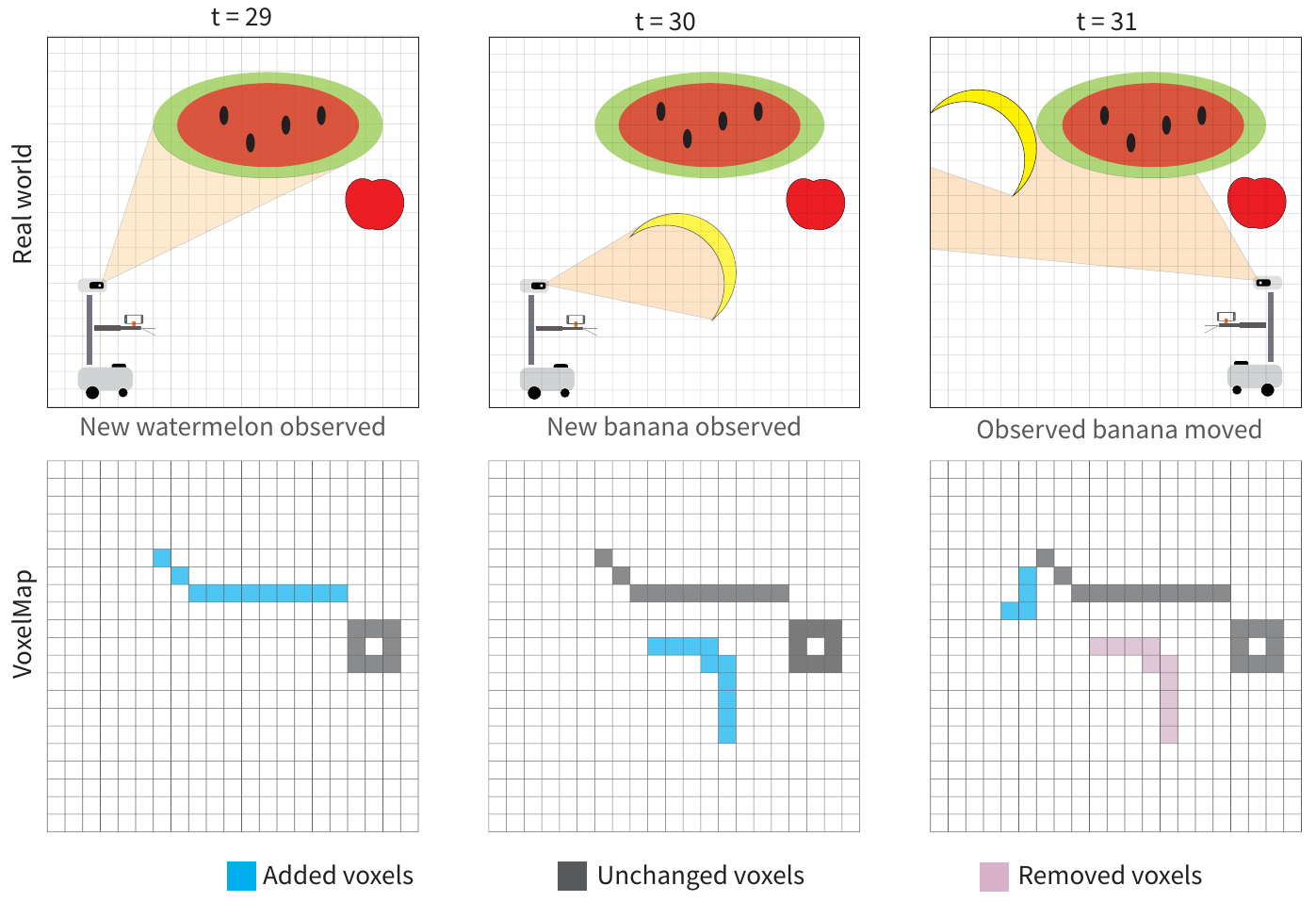}
    \caption{A high-level, 2D depiction of how adding and removing voxels from the voxel map works. New voxels are included which are in the RGB-D cameras view frustum, and old voxels that should block the view frustum but does not are removed from the map.}
    \label{fig:voxelmap-add-delete}
\end{figure}
We illustrate a simplified 2D representation of this algorithm in Figure~\ref{fig:voxelmap-add-delete}. In practice, to speed up the intersection between the sparse voxelmap and the view frustum, we project each existing voxel to the camera plane and calculate the camera distance. If the image height and width are $(H, W)$, the depth image is $\mathbf{D}$, and a certain voxel is projected to points $(h, w)$ in the camera plane with depth $d$, it gets removed if both Eq.~\ref{eq:camera-plane} and~\ref{eq:depth-plane} hold.
\begin{align}
    (h, w) & \in \left [0, H\right ] \times \left [0, W\right ]  \label{eq:camera-plane}                 \\
    d      & \in \left ( 0, \min(2, \mathbf{D}\big[h, w\big] + \epsilon) \right ) \label{eq:depth-plane}
\end{align}
Where Eq.~\ref{eq:camera-plane} ensures that the point falls within the camera view, and Eq~\ref{eq:depth-plane} ensures that (a) the depth $d > 0$, or the object is in front of camera, (b) $d < 2$m, or the voxel isn't too far away from the camera, and (c) $d < \mathbf{D}[h, w]$ denoting the voxel is between the camera and the currently visible object.

\subsection{Querying~\MODEL{} for Object Localization}
\label{subsec:grounding}

As described in Section~\ref{subsec:problem}, we define the object localization or 3D visual grounding problem as a function mapping a text query and posed RGBD images to either the 3D coordinate of the query object, or $\emptyset$ if the object is not in the scene.
Unlike previous work, we abstain from returning a location when an object is not found.
To enable this, we factor this grounding problem into two sub-problems.
The first is finding the latest image where the queried object could have appeared.
The second is identifying whether the object is actually present in that image.
For the first sub-problem, we introduce two alternate approaches of visual grounding: one using the intrinsic semantic features of~\MODEL{}, and another using state-of-the-art multimodal LLMs such as GPT-4o~\cite{gpt4} and Gemini 1.5 Pro~\cite{gemini}.

\mysection{Embedded Vision Language Features}
\label{subsec:vlm-features}
Vision Language Models (VLMs) such as CLIP~\cite{clip} and SigLIP~\cite{siglip} possess an ability to embed both images and languages into the same latent space, where the similarity between an image and a text object can be calculated by simply taking the dot product between the two latent representation vectors.
We use this property of the embedding vectors to query our voxel map with open-vocabulary text queries.
\begin{figure}
    \centering
    \includegraphics[width=\linewidth]{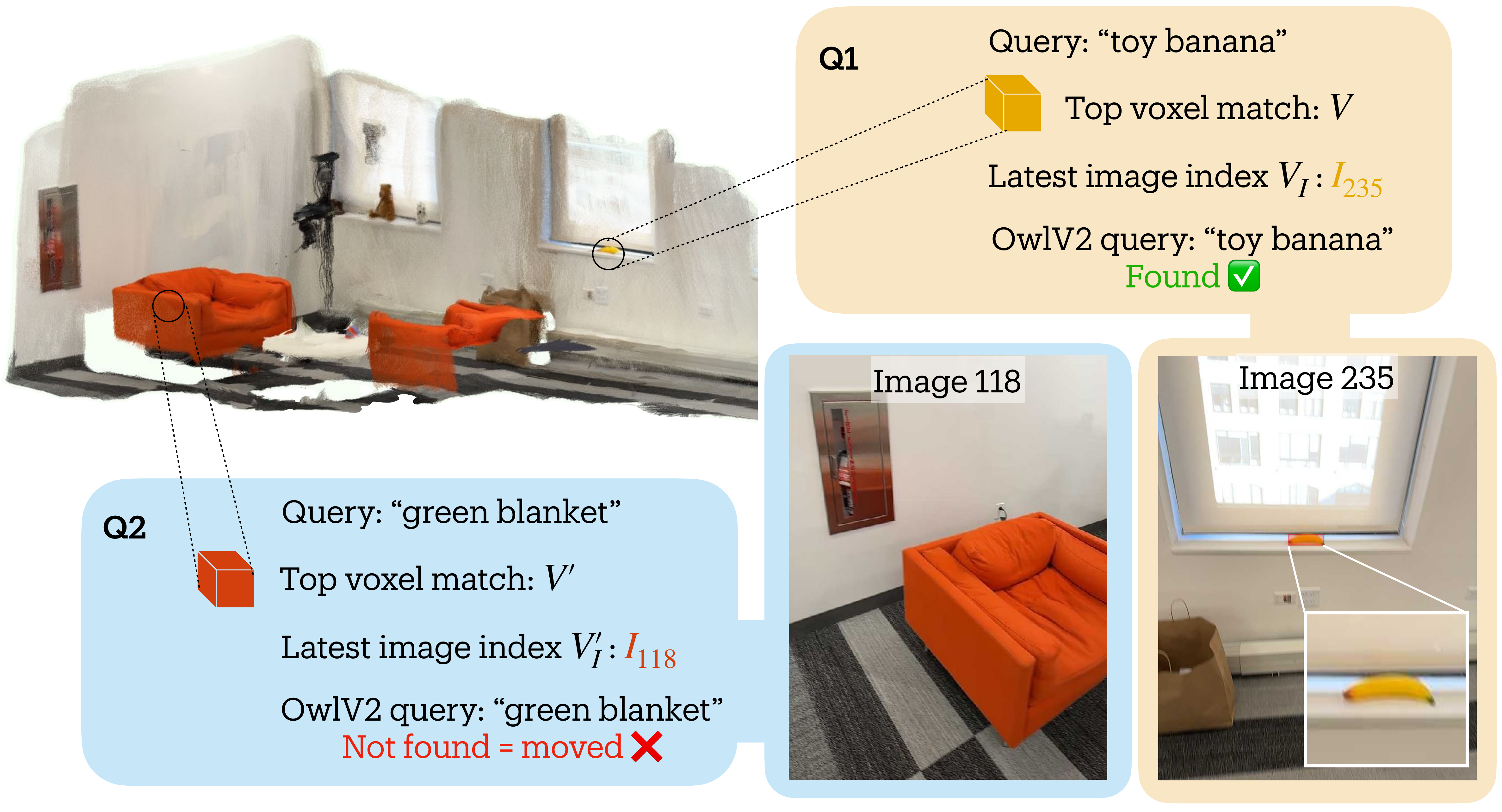}
    \caption{Querying~\MODEL{} with a natural language query. First, we find the voxel with the highest alighnment to the query. Next, we find the latest image of that voxel, and query with an open-vocabulary object detector to confirm the object location or abstain.}
    \label{fig:dynamem-object-query}
\end{figure}
As described in Section~\ref{subsec:updating}, we convert the incoming images to point-wise image features, and embed them into our voxels.
When we have a new language query, we calculate its latent embedding using the VLM text encoder, and find the voxel whose feature has the highest dot product with the text embedding.
Once we find the right voxel, we simply retrieve its associated latest image from our data structure as shown in Figure~\ref{fig:dynamem-object-query}.

As a bonus feature, our VoxelMap can also return $k>1$ possible objects and associated images for a single query. We do this by using a DBSCAN clustering of voxels similar to~\cite{llmgrounder}, and returning the images associated with the most aligned voxel in top-$k$ clusters.

\mysection{Multimodal Large Language Models (mLLMs)}
\label{subsec:mllm-features}
For this approach, we note that the problem of finding the latest image where an object may appear is similar to the problem of visual question-answer (VQA)~\cite{antol2015vqa}. Since we fully rely on pretrained models to build our map, we pose this multi-image VQA problem as an mLLM QA problem similar to OpenEQA~\cite{majumdar2024openeqa}.

\begin{figure}
    \centering
    \includegraphics[width=\linewidth]{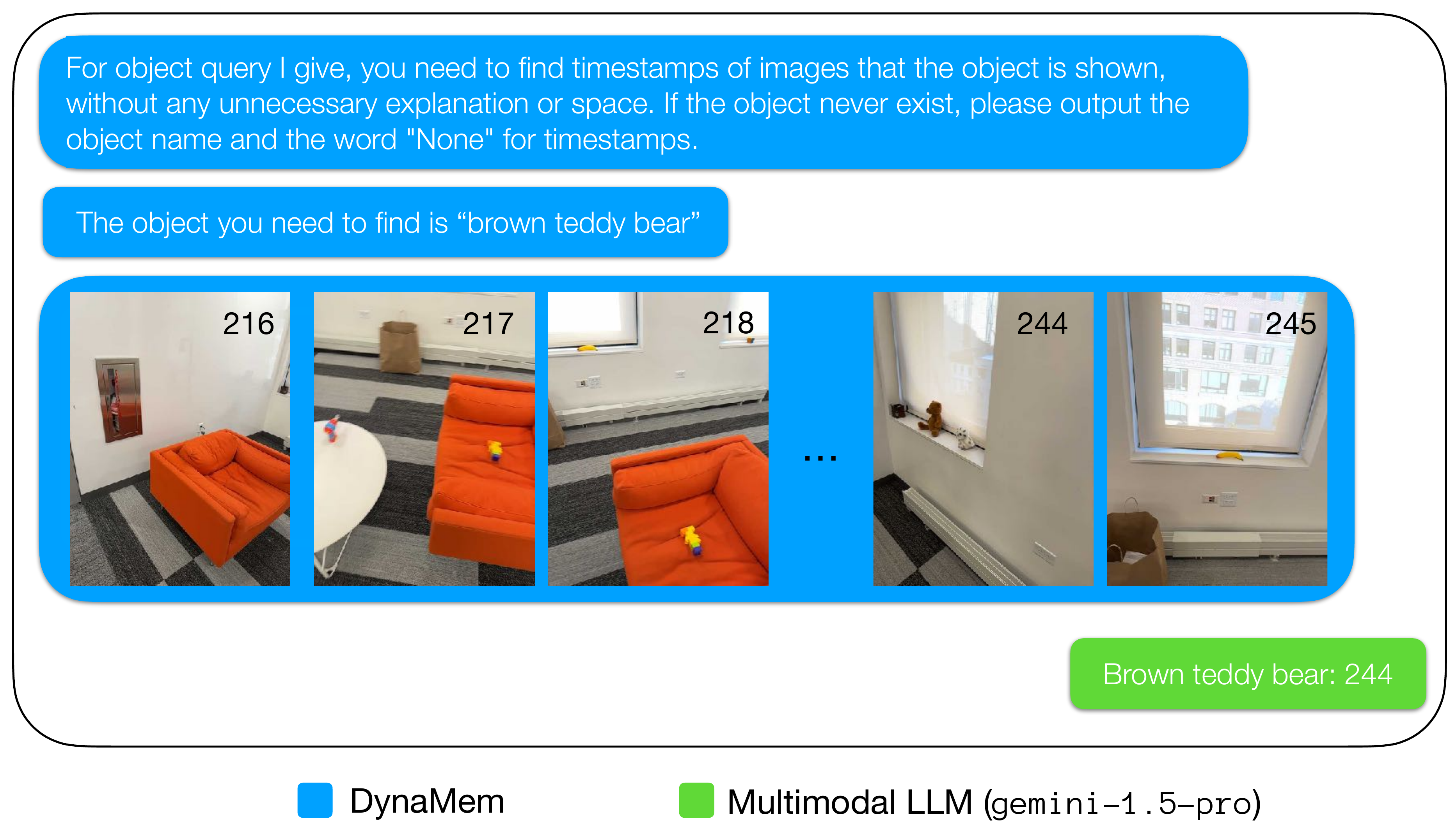}
    \caption{The prompting system for querying multimodal LLMs such as GPT-4o or Gemini-1.5 for the image index for an object query.}
    \label{fig:mllm-query}
\end{figure}
We show in Figure~\ref{fig:mllm-query} how we query the mLLMs to solve the visual grounding query.
We give the model a sequence of our latest environment observations images and ask the model for the index of the last image where the queried object was observed.
We additionally instruct the model to respond ``None'' if the object was not observed in any image.
Note that, unlike OpenEQA~\cite{majumdar2024openeqa}, we only pass the RGB images to the mLLM, and not the depth or camera pose.
Similarly, we only ask for an image index, and not a full textual answer.

One important hyperparameter for this mLLM query method is the maximum number of images included in the prompt. Longer context needs longer processing time and potentially includes outdated information, while short context might not include all information and thus will miss objects. We optimize the context by excluding completely outdated images: all images $I$ with no voxel pointing to them are deleted. This filtering increases mLLM context utilization. %
We set Gemini as our base model and 60 as our query image limit since Gemini context can fit 60 images, which is twice as large as GPT-4o's context size.

\mysection{Combining the Approaches}
\label{subsec:combining}
From the discussion above, and from our real-world experiments as shown in Section~\ref{sec:experiments}, we see that the downsides of these two methods in practice are somewhat complementary. The VoxelMap can easily process a large number of observation images over time, but it struggles to disambiguate between multiple similar but not quite same objects. On the other hand, mLLM based methods can distinguish between fine differences in query objects, but can only handle few images at a time. As a result, we come up with a \textit{hybrid} approach -- taking advantage of the best of both methods.

For this approach, we build and process the VoxelMap as usual. For the hybrid querying, parametrized by an integer $k$, we retrieve top $k$ candidate images from the VoxelMap for the given query. Then, we pass those $k$ images into the mLLM, and query them as usual for the mLLM approach. The mLLM answers with the latest image where the query object may appear, which we then use for the downstream processing. Note that, this approach generalizes both of our individual approaches: when $k = 1$, it converges to the VLM-feature-only approach, and when $k \rightarrow \infty$ it converges to the mLLM-only approach.

\mysection{Handling Absence of Queried Object}
\label{subsec:abstaining}
Several previous methods~\cite{clipfields,lerf,okrobot} assume that the queried object is always present in the scene, and always responds with the object that is the best match to the query. However, this often results in high false-positive failure cases. For example, in a scene with no red cups and a blue cup, the method may respond with the location of the blue cup in response to the query ``red cup''.

For this reason, we locate objects in two stages. First, we find the best candidate image where the object may have been seen (Section~\ref{subsec:vlm-features}).
Then, we use an open-vocabulary object detector model such as OWL-v2~\cite{owlv2} to search that image for the queried object (Figure~\ref{fig:dynamem-object-query}).
If we don't find the queried object, we assume that the object has either moved, or the response from the voxelmap or mLLM was inaccurate, and respond with ``object not found''.
If the open-vocabulary object detector returns an object bounding box, we find the median pixel from the object mask and return its 3D location. %

\subsection{Robot Navigation and Exploration}
\label{subsec:path_planning}
To navigate in a real-world environment, robots use an obstacle map in conjunction with a navigation algorithm like A* in~\cite{vlmaps, okrobot}.
We use a simple voxel-projection strategy to build an obstacle map. Due to the depth observation noise, we simply set a threshold for the ground (0.2m for our experiments), and project all the voxels above that $z$-threshold as the obstacles in our map.
The voxels below the threshold are projected into the 2D obstacle map as navigable points.
Finally, the points in the map that are not marked as either obstacle or navigable are marked as explorable points.

\mysection{Exploration Primitives}
Since our robot does not start with an environment map, it explores the environment with frontier based methods to build the map.
We can further accelerate this process by providing exploration guidance.
Based on the current status of the map,~\MODEL{} provides an exploration value function to accelerate the exploration process both for building and updating the map.

We provide two value-based exploration maps: one time-based, and one semantic-similarity-based~\cite{yokoyama2024vlfm}. The time-based value map prioritizes the least-recently seen points. If the current time is $T$, and the last-seen time of voxel $(x, y, z)$ is $t_{x, y, z}$, the temporal value map $\mathbb{V}_T$ is expressed as:
\begin{align*}
     & \mathbf{T}^*[x, y] = \max_z (T - t_{x, y, z})                                        \\
     & \mathbb{V}_T[x, y] = \sigma\left(-\beta_T\big(\mathbf{T}^*[x, y] - \mu_T\big)\right)
\end{align*}
where $\beta_T, \mu_T$ are hyper-parameters and $\sigma$ is the sigmoid function. Similarly, if the VLM feature at voxel $(x, y, z)$ is $f_{x, y, z}$, and the VLM feature for the language query is $f_q$, then the similarity-based value map $\mathbb{V}_S$ is be expressed as:
\begin{align*}
     & \mathbf{S}^*[x, y] = \max_z (f_q \cdot f_{x, y, z})                                  \\
     & \mathbb{V}_S[x, y] = \sigma\left(-\beta_S\big(\mathbf{S}^*[x, y] - \mu_S\big)\right)
\end{align*}
where once again $\beta_S, \mu_S$ are hyperparameters. We may also linearly combine $\mathbb{V}_T, \mathbb{V}_S$ to balance our exploration between last seen time and semantic similarity.

Finally, since the environment may be dynamic, we convert our navigation algorithm from open-loop to closed-loop. The robot, instead of executing the entire navigation plan generated by A*, stops after the first seven waypoints (approx. 0.7 to 1 meters). Then, the robot scans the environment, updates the map, and moves according to a new plan. The robot repeats these steps until its distance to the target is lower than a predefined threshold.

\section{Experiments}
\label{sec:experiments}
We evaluate our method,~\MODEL{}, on a Hello Robot: Stretch SE3 in real world environments, and perform a series of ablation experiments in an offline benchmark.

\begin{figure}[htb!]
    \centering
    \includegraphics[width=\textwidth]{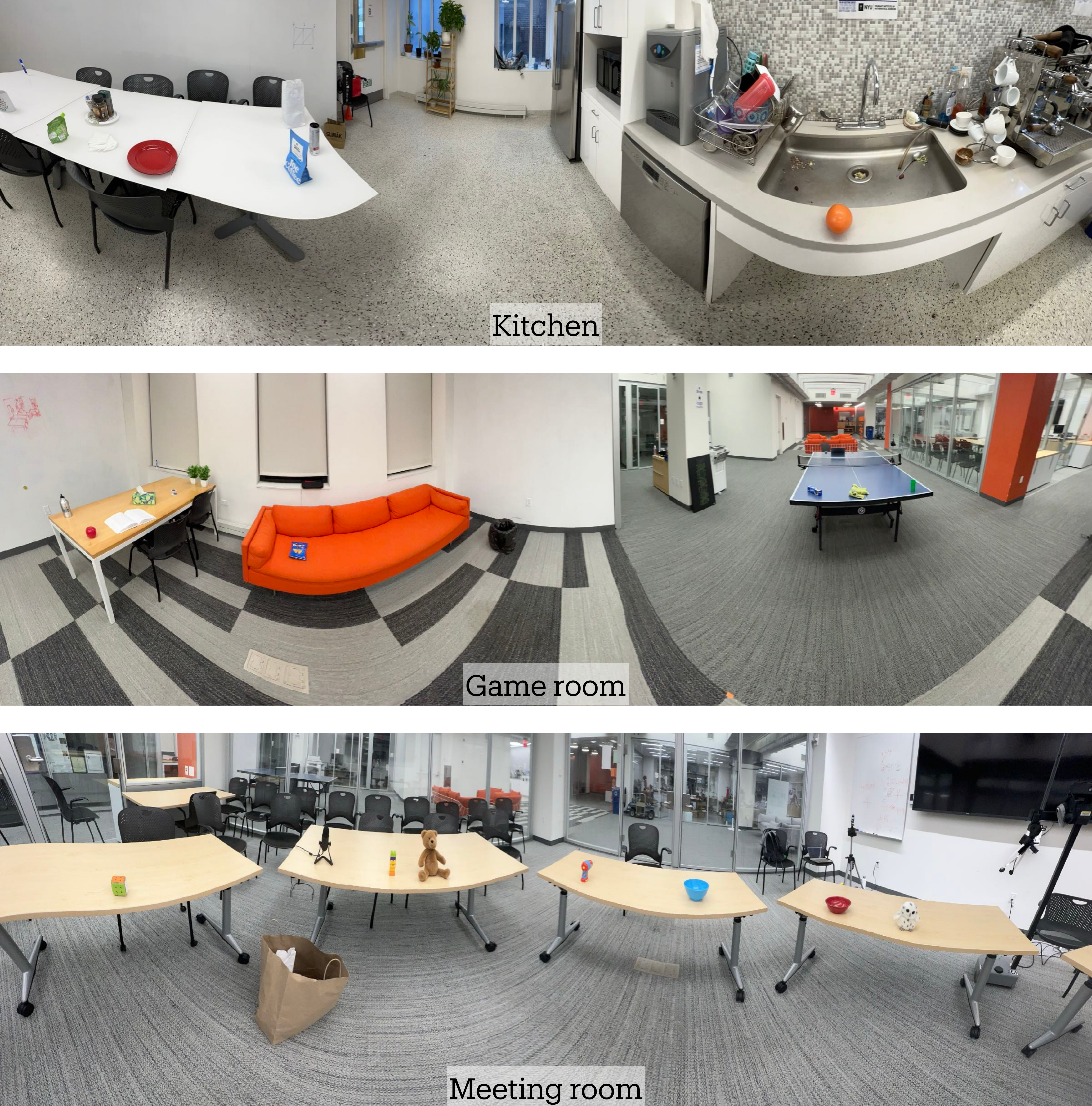}
    \caption{Real robot experiments in three different environments: kitchen, game room, and meeting room. In each environment, we modify the environment thrice and run 10 pick-and-drop queries.}
    \label{fig:env_examples}
\end{figure}

\subsection{Real-world Experiments}
\label{sec:real-world-experiments}
We evaluate~\MODEL{} and its impact on open-vocabulary mobile manipulation in three real-world dynamic environments (Figure~\ref{fig:env_examples}).
In each environment, we set up multiple objects as potential manipulation targets, change the environment in  three rounds, and execute 10 pick-and-drop queries over the rounds
We use the Hello Stretch SE3 as our mobile robot platform, and use its head-mounted Intel RealSense D435 RGB-D camera to collect the input data.

To build a complete pick-and-drop system around~\MODEL{}, we follow the system architecture in OK-Robot~\cite{okrobot}.
In particular, we use the AnyGrasp~\cite{fang2023anygrasp} based open-vocabulary grasp system and use the heuristic based dropping system.
However, we use~\MODEL{}'s exploration primitives let the robot build the map of the environment and allow the robot to explore when an object is not found in the memory.

As a baseline, we compare with OK-Robot~\cite{okrobot}, a state-of-the-art method for OVMM.
OK-Robot uses a static voxelmap as its memory representation, and thus it highlights the importance of dynamic memory for OVMM in a changing environment.
For~\MODEL{}, we run two variations of the algorithm in the real world: one with VLM-feature based queries and one with mLLM-QA based queries.

\begin{figure}[thb]
    \centering
    \includegraphics[width=\textwidth]{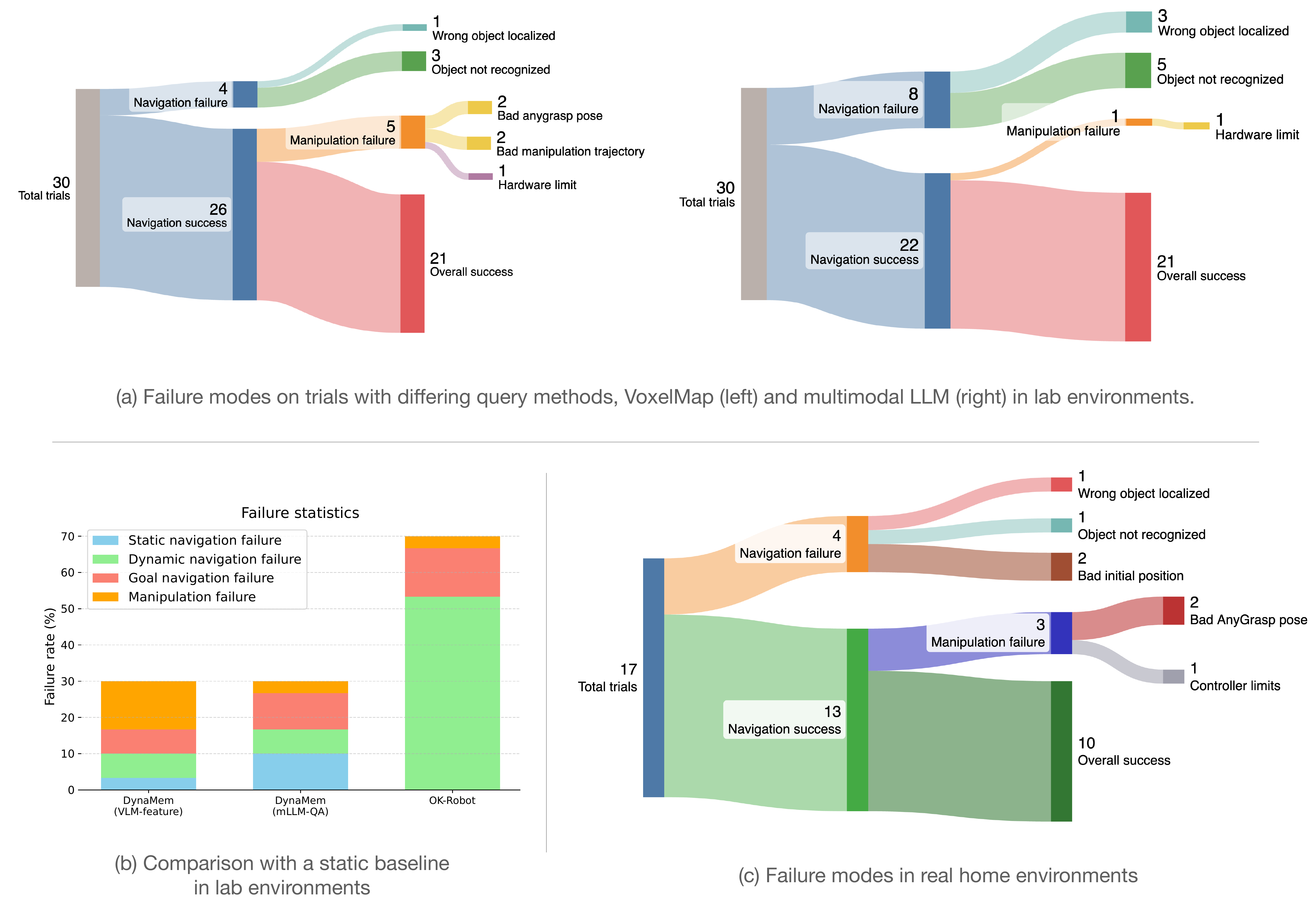}
    \caption{Statistics of failure, broken down by failure modes, in our real robot experiments in the lab and in home environments. Statistics are collected over three environments and 30 open-vocabulary pick-and-drop queries for the lab experiments, and two environments and 17 pick-and-drop queries for the home environments, on objects whose locations change over time.}
    \label{fig:failure}
\end{figure}
\mysection{Results}
\label{sec:real-world-results}
Our experiments in three dynamic environments and with 30 queries is summarized in Figure~\ref{fig:failure}(a \& b). We find that~\MODEL{} with both VLM-feature based and mLLM-QA based queries have a total success rate of 70\%. This is a significant improvement over the OK-Robot system, which has a total success rate of 30\%.
Notably,~\MODEL{} is particularly adept at handling dynamic objects in the environment: only 6.7\% of the trials failed due to our system not being able to navigate to such dynamic objects in the scene.
This is in contrast to the OK-Robot system, where 53.3\% of the trials failed because it could not find an object that moved in the environment.
In contrast, navigating to static goals fails in only 10\% of the cases for~\MODEL{} with VLM-feature, 13.3\% for OK-Robot and 20\% for~\MODEL{} with mLLM-QA.

We observe that a common localization failure for VLM-feature based queries is that those features often act like a bag-of-words model; for example they may find a blue bowl when queried for a red bowl. On the other hand, mLLM-QA based queries often fail because the objects are not present in the short context window, or the images in the context window are outdated.
Since their failure modes are complementary, we hope to combine the two methods in future work to improve the overall performance.
\subsection{Deployment in Home Environments}
\label{sec:home-exp}
To understand the applicability of our system beyond lab environments that emulate real living space, we deployed~\MODEL{} to two apartments in New York City. We ran experiments with~\MODEL{} on a one one-room and one two-room environment. We ran a total of 17 long-horizon trials to understand the possible failure modes when deployed on a real scene. Since we observe complementary cases of failure from the two querying methods (Section~\ref{subsec:combining}) in our lab experiments in Section~\ref{sec:real-world-results}, we run these experiments with the hybrid querying method, setting the number of images returned by VoxelMap to be processed by mLLMs to be $k=3$.

We found a total of 9 successes in our 17 real home trials, but it is significantly more interesting for us to understand how these systems may fail in real environments, as shown in Figure~\ref{fig:failure}(c). We first observe that out of the 8 failures, 4 happened due to poor navigation, 3 happened due to our manipulation skills failing, and 1 was an error in placing the object at the target. Out of the navigation failures, half happened because of the target object was not found or was confused with another object, and the other half happened due to navigating to a suboptimal location for manipulation.

\subsection{Ablations on an Offline Benchmark}
\label{sec: benchmark}
Running real robot OVMM experiments can be expensive and time-consuming.
So, we developed an offline benchmark called~\BENCH{} to easily evaluate dynamic 3D visual grounding algorithms on dynamic environments and perform algorithmic ablations.
The benchmark isolates the query-response part of the dynamic semantic memory without robot navigation, exploration, and manipulation.

\mysection{Data Collection}
\label{subsec: data_collection}
In the real world, the robot collects its own map-building data by exploring the environments.
Following this, we collect the robot's runtime sensor data from three environments.
To further enrich our benchmark, we simulate this process by taking posed RGB-D images on an iPhone Pro in six more environments.
In all cases, we emulate environment dynamics by moving objects and obstacle locations in three successive rounds.

\mysection{Data Labeling and Evaluation}
\label{subsec: data_labelling}
We manually annotate queries and responses in the dataset.
Each query has an associated natural language label $q$, object location $\vec{X} =(x, y, z)$, and an object radius $\epsilon$.
Since the environment is dynamic, each query also has an associated time $t$.
For evaluation, at time $t$ (i.e. after the memory algorithm has observed all the input data points with timestamp $< t$), we query the model with $q$.
If it predicts an object location $\vec{X'} = (x', y', z')$, it's a success if $||X-X'||_2 \le \epsilon$ and a failure otherwise. Since the robot may also encounter queries for objects it has not observed yet, we emulate negative queries by adding queries for objects (a) that have not been observed yet, or (b) that have been observed but were subsequently removed. For both of these query types, the model must respond with \textit{not found}; otherwise it's counted as a failure.

\mysection{Evaluation Results}
\label{sec: benchmark-eval}
Using our offline benchmark, we ablate design decisions of~\MODEL{} as discussed in Section \ref{sec:method}.
Among these design decisions, the primary are: using feature embedding-based vs. mLLM-QA based language grounding, ablating components such as point removal or abstentiation from the algorithm, and trying different mLLMs.
Due to API costs, we only evaluate Gemini models on the benchmark.
We present our results in Table~\ref{table:benchmark}.

\begin{table}[tb!]
    \caption{Ablating the design choices for our query methods for \MODEL{} on the offline \BENCH{} benchmark. We also present results from five human participants to ground the performances.
    }
    \centering
        \begin{tabular}{@{}llc@{}}
            \toprule
            Query type  & Variant                               & \multicolumn{1}{l}{Success rate} \\ \midrule
            Human       & (average over five participants)         & \textbf{81.9\%}                  \\ \midrule
            VLM-feature & default (adding and removing points)  & \textbf{70.6\%}                  \\
                        & only adding points                    & 67.8\%                           \\
                        & no OWL-v2 cross-check                 & 59.2\%                           \\
                        & no similarity thresholding            & 66.8\%                           \\ \midrule
            mLLM-QA     & default (Gemini Pro 1.5)              & \textbf{67.3\%}                  \\
                        & Gemini Pro 1.5, no voxelmap filtering & 66.8\%                           \\
                        & Gemini Flash 1.5                      & 63.5\%                           \\\midrule
            Hybrid      & VLM-feature $\rightarrow$ mLLM $(k=3)$                    & \textbf{74.5\%}                  \\
            \bottomrule
        \end{tabular}%
        \label{table:benchmark}
\end{table}
We see that performance of  VLM-features and mLLM-QA follows the same order in the real world in the benchmark, corroborating the benchmark design.
The best design choices are to both add and remove points, and to cross check with OWL-v2 on top of similarity thresholding for VLM-feature based grounding. 
For mLLM-QA based grounding, Gemini Pro outperforms Gemini Flash, and voxelmap based image filtering benefits the method.
Moreover, we see that the hybrid method that uses VoxelMap feature filtering and then sends the top 3 images to the mLLM performs better than either method individually.

\section{Conclusions and Limitations}

In this work, we introduced~\MODEL{}, a spatio-semantic memory for open-vocabulary mobile manipulation that can handle changes to the environment during operation.
We showed in three real world environments that~\MODEL{} can navigate to, pick, and drop objects even while object and obstacle locations are changing.
In the future, we could improve~\MODEL{} performance by merging the VLM-feature queries and mLLM-QA queries, as they show complementary failure cases.
Similarly, reasoning over both object and voxel level abstraction could speed up environment update when objects move.
Our current system experiences a large number of manipulation failures: using mLLMs to detect and recover from failures~\cite{etukuru2024robot} may increase the performance of the overall system.
Finally, integrating with more skills, such as searching in cabinets or drawers, would improve the applicability of such OVMM systems in the real world.

\bibliography{references}

\end{document}